\documentclass[journal]{IEEEtran}
\usepackage{amsmath}
\usepackage{newtxtext,newtxmath}
\usepackage{booktabs}
\usepackage{graphicx}
\usepackage{multirow}
\usepackage{xcolor}
\usepackage{placeins}
\usepackage[hidelinks]{hyperref}

\newcommand{\note}[1]{}
\newcommand{\repourl}{\url{https://github.com/JialingRichard/nextunit-bench}}

\begin{document}

\title{Moving Generality from Model to Modeler:\\
LLM Agents for PHM on a Deployment-Style Benchmark}

\author{Zhenling~Chen, Zhiguo~Zeng, and Anne~Barros%
\thanks{Z.~Chen is with Laboratoire G\'enie Industriel, CentraleSup\'elec, Universit\'e Paris-Saclay, Gif-sur-Yvette, France (e-mail: zhenling.chen@centralesupelec.fr).}%
\thanks{(Corresponding author: Zhiguo Zeng.)}%
\thanks{Z.~Zeng and A.~Barros are with the Chair on Risk and Resilience of Complex Systems, Laboratoire G\'enie Industriel, CentraleSup\'elec, Universit\'e Paris-Saclay, Gif-sur-Yvette, France (e-mail: zhiguo.zeng@centralesupelec.fr; anne.barros@centralesupelec.fr).}}

\maketitle

\begin{abstract}
Coding agents built on large language models (LLMs) now help develop many models published as general solutions. Should generality instead reside in the agent that builds a model for each new task? We examine this question in prognostics and health management (PHM), where every model must work on unseen equipment. Our deployment-style benchmark, NextUnit-Bench, covers 24 public datasets split into 58 folds that hold out units or operating conditions; each agent has two hours to deliver an executable predictor, then run on the hidden units. First, frontier agents outperform AutoML, which reaches less than half the best agent's score, and faithful implementations of published methods on four datasets, of which only 6 of 30 beat the strongest agent's own design; published accuracies, where reproducible, rest on the papers' own conventions, and a single convention, capping the remaining life of test engines, can halve the reported error. Second, the agents' advantage is a design made on each machine's data; once made, a plain-text plan of a few hundred words carries most of the gap between two frontier agents, and given a weaker agent's plan, the strongest agent falls to that agent's level, often to identical predictions. Before touching the data, however, it matters little whose plan an agent carries out, and domain toolkits appear to help neither: what sets the strongest apart is how much it learns from the data. Third, such designs reach only implementers that can execute them: given the strongest agent's designs, a locally deployable open-weight model delivers half as often, even when they are lightened, and scores no higher than without them. Generality, we argue, is not disappearing but moving, from the model to the modeler.
\end{abstract}

\begin{IEEEkeywords}
Prognostics and health management, LLM agents, benchmark, reproducibility, knowledge transfer, deployment.
\end{IEEEkeywords}

\section{Introduction}

Coding agents built on large language models (LLMs) have quietly become part of how models are developed in science and engineering. Researchers and engineers use them to build models faster, and often better, and then publish the resulting, seemingly superior general models as papers. This practice raises two questions. The first is about credit: in such a development process, how much of the contribution comes from the human developer, and how much from the agent? The second is about generality: once the contribution is published as a general model or algorithm, how much of its performance survives on a task that is similar but not the same? Taken together, they point to a more radical question. If a state-of-the-art agent can already build, for each new task, a model as good as a human developer's, and do so faster, should generality reside in the agent, the modeler, rather than in an ever-growing stream of narrow, task-specific models?

Prognostics and health management (PHM) is a natural place to put this question to the test, because every PHM model eventually meets a machine it has never seen: the next bearing, an engine that has not yet failed, an air-handling unit in another building. A general model earns its name only if it keeps working there. We therefore examine agents the way models are actually used. An agent receives a task and the data of a set of machines, builds a model on its own within two hours, and delivers an executable predictor, which we then run on machines it has never seen. On 24 public PHM datasets, we pit frontier and locally deployable agents against faithful implementations of published methods and against AutoML.

The first result surprised us: the agents win. Published methods, implemented faithfully, and AutoML pipelines both fall behind what a general-purpose agent builds for the equipment at hand; of 30 implementations of 20 published methods on four datasets, only 6 beat the strongest agent's own design. If an agent can out-model the literature on equipment it has only just met, where does its advantage come from? Is it a better programmer, or does it make better decisions?

To find out, we write down the design an agent delivered as a plain-text plan of a few hundred words, with no code, and hand it to another agent to implement from scratch. The answer is striking. Given a weaker agent's plan, the strongest agent becomes the weaker agent: its scores fall to the weaker agent's level, often down to identical predictions, even though it writes every line of code itself. Given the strongest agent's plan, the weaker agent moves toward it, dataset by dataset. The advantage is a design made for this particular machine, tried on its data down to the hyperparameters; a published method is such a design too, only chosen for other machines. Little of it lies in heavy machinery: stripped of neural networks, ensembles, and specialized packages, the strongest agent's designs keep most of their value. Could the strongest agent have written those words without the data? Asked to plan each task from its description, its documentation, and the layout of its files, without seeing a single value, its plans were worth only about a point and a half more than the weaker agent's to the same implementer, against 6 to 13 points for plans written on the data, and toolkits of domain knowledge appeared to help neither. What sets the strongest agent apart is not what it knows but how much it learns from the data in front of it.

These few hundred words, however, reach only those who can carry them out. Handed the strongest agent's plans, a locally deployable open-weight model delivers a working predictor only half as often as on its own and ends up no better than with no plan at all; even lightened, the plans remain beyond it. Yet what it does deliver is no worse than what it builds on its own. What a local model lacks is not knowledge, but the capacity to act on it. And because the value is made on the data, which in many plants may not leave the site, what local deployment needs is a capable modeler beside the data, not more knowledge from outside.

These findings bear on a question that PHM has rarely asked. Transfer learning and domain adaptation are among the most popular topics in the field, and generalization receives a great deal of attention, yet we seldom ask what we really need it for. Is it to improve accuracy, or to save cost? The history of the field hints at an answer. PHM moved from signal processing to machine learning and then to deep learning, and each step promised to spare experienced engineers the work of hand-crafting features for every machine. Had there been countless such engineers, enough to tailor feature extraction to every machine, would this progression have happened at all?

LLM agents are beginning to make that counterfactual real. Systems built on LLMs and agents keep getting cheaper, while their accuracy and capability keep growing, so much so that we can imagine an agent writing custom models and algorithms for every individual physical asset. Do we then still need generalization in the traditional sense? Our results suggest that generality is not disappearing but moving: from a model expected to work everywhere to a modeler that builds the right model for each machine. We make this case through four contributions:
\begin{enumerate}
\item NextUnit-Bench, an open deployment-style benchmark (\repourl) that evaluates agents the way models are used: built on one set of machines, delivered as a program, and run on machines never seen;
\item evidence that, on this benchmark, frontier agents outperform AutoML averaged over 24 datasets, that the strongest of them outperforms faithful implementations of published methods on average on three of four datasets, though the best of those methods beat it on two, and that published accuracies depend on the papers' evaluation conventions;
\item plan transplant and plans written without the data, which show that the agents' advantage is a design made on each machine's data that, once made, fits in a plain-text plan of a few hundred words and travels between agents able to execute it; that before touching the data, it matters little whose plan an agent carries out; and that neither heavy machinery nor domain toolkits are its source;
\item a diagnosis of locally deployable agents, which struggle to execute such designs even when they are lightened, yet build them no worse when they do: capability buys the probability of completing a design, not its quality.
\end{enumerate}

\section{Related Work}

\subsection{LLM Agents for Machine-Learning Engineering}
LLM agents that build machine-learning pipelines end to end are now measured on benchmarks drawn from Kaggle competitions, such as MLE-bench~\cite{chan2024mlebench}, and built on scaffolds that search over candidate code, such as AIDE~\cite{jiang2025aide}. MLE-Dojo~\cite{qiang2025mledojo} provides interactive environments for training such agents, TimeSeriesGym~\cite{cai2025timeseriesgym} extends the setting to time-series tasks, and recent work finds that, with a strong model, a minimal single-session coding agent does as well as elaborate scaffolds~\cite{harness2026}, which is why we use plain command-line agents. These benchmarks score predictions on fixed splits of competition-style data. None of them asks whether a delivered pipeline keeps working on physical units it has never seen, the question on which PHM stands or falls.

\subsection{LLMs and Agents in PHM}
In PHM, LLMs have so far been evaluated mostly as assistants rather than as modelers. PHM-Bench~\cite{phmbench2025} scores large models on domain knowledge, algorithm generation, and task optimization across PHM tasks; PHMForge~\cite{phmforge2026} and AssetOpsBench~\cite{patel2026assetopsbenchkdd} have agents call tools that wrap published algorithms or orchestrate agents for asset operations; agents have been asked to diagnose a simulated appliance step by step from its sensors~\cite{coffeebench2026}; LLMs placed above existing diagnostic tools fail to exploit them consistently~\cite{kim2026llmtools}; and a multi-agent framework fuses bearing-domain knowledge into an LLM's reasoning and selects feature pipelines by signal characteristics, without updating model weights~\cite{diagent2026}. A recent review discusses foundation models for PHM~\cite{han2026fmphm}. In all of these, the models themselves are built by humans or provided as tools. In ours, the agent builds the model from scratch, delivers it, and is judged by how it runs on equipment it has never seen.

\subsection{Reproduction and Evaluation Conventions in PHM}
Evaluation conventions are known to shape reported results in PHM. On CWRU, segments cut from the same recordings end up in both training and test sets, which inflates accuracy~\cite{hendriks2022cwru}, and remaining-useful-life results on C-MAPSS~\cite{saxena2008cmapss} are customarily reported against capped labels. Agents have recently been used to reproduce under-specified methods in machine health intelligence~\cite{theiler2026paper}. We go one step further and let the same agent implement each published method twice, once under the paper's own protocol and once under deployment, which separates the effect of the convention from that of the implementation.

\section{A Deployment-Style Benchmark}
Our benchmark, NextUnit-Bench (NextUnit for short), asks one question of every method: given the data of some machines, does what you deliver work on machines you have never seen?

\subsection{Tasks and Folds}
The benchmark comprises 24 public datasets (Table~\ref{tab:datasets}) in four families: condition classification, remaining useful life (RUL), degradation and health estimation, and anomaly detection or failure prediction. The assets range from bearings, gearboxes, and milling cutters to battery cells, turbofan engines, wind turbines, oil wells, and trucks. Every dataset is split by physical unit, such as a bearing, a cell, a cutter, an engine, or a well, so that no test unit is seen during development; where a dataset has no independent units, as for CWRU, SEU, and the UCI hydraulic rig, whole operating conditions are held out instead, and official splits are kept where they exist. This yields 58 folds.

\begin{table*}[t]
\centering
\caption{The 24 datasets of the benchmark, grouped by task family.}
\label{tab:datasets}
\footnotesize
\begin{tabular}{lllrrll}
\toprule
Dataset & Asset & Task & Units & Folds & Mode & Metric \\
\midrule
\multicolumn{7}{l}{\emph{Condition classification}}\\
Paderborn~\cite{lessmeier2016paderborn} & Rolling bearing & 4-class condition & 32 & 3 & Batch & Accuracy \\
CWRU~\cite{smith2015cwru} & Rolling bearing & 4-class fault & 40$^{a}$ & 4 & Batch & Accuracy \\
SEU~\cite{shao2019highly} & Gearbox & 5-class fault & 10$^{a}$ & 2 & Batch & Accuracy \\
UCI hydraulic~\cite{helwig2015hydraulic} & Hydraulic rig & Valve condition & 3$^{a}$ & 3 & Batch & Balanced accuracy \\
Tennessee Eastman~\cite{rieth2017tep} & Chemical process$^{b}$ & Fault class & 2100 & 1 & Batch & Accuracy \\
Petrobras 3W~\cite{vaz2026threew} & Oil well & Event class & 40 & 3 & Batch & Balanced accuracy \\
AHU~\cite{nehasil2021ahu} & Air-handling unit & Fault detection & 6 & 3 & Batch & Balanced accuracy \\
SCANIA Component X~\cite{kharazian2025scania} & Truck & Time-to-repair class & 40$^{c}$ & 1 & Batch & Balanced accuracy \\
\midrule
\multicolumn{7}{l}{\emph{Remaining useful life}}\\
XJTU-SY~\cite{wang2020xjtu} & Rolling bearing & RUL & 15 & 3 & Stream & MAE \\
PRONOSTIA~\cite{nectoux2012pronostia} & Rolling bearing & RUL & 17 & 1 & Batch & MAE \\
IMS~\cite{qiu2006ims} & Rolling bearing & RUL & 4 & 3 & Stream & MAE \\
C-MAPSS FD004~\cite{saxena2008cmapss} & Turbofan engine$^{b}$ & RUL & 497 & 1 & Batch & RMSE \\
N-CMAPSS DS02~\cite{arias2021ncmapss} & Turbofan engine$^{b}$ & RUL & 9 & 1 & Stream & RMSE \\
PHM Europe 2020~\cite{phme2020challenge} & Filter & RUL & 32 & 3 & Stream & MAE \\
MATR~\cite{severson2019matr} & Li-ion cell & Cycle life & 124 & 1 & One-shot & MAE \\
\midrule
\multicolumn{7}{l}{\emph{Degradation and health estimation}}\\
MATWI~\cite{depauw2023matwi} & Milling cutter & Tool wear & 17 & 3 & Batch & MAE \\
PHM 2010~\cite{phm2010challenge} & Milling cutter & Flank wear & 3 & 3 & Stream & MAE \\
NASA randomized battery~\cite{bole2014randomized} & Li-ion cell & Capacity & 28 & 3 & Stream & MAE \\
NASA battery~\cite{saha2007battery} & Li-ion cell & Capacity & 33 & 3 & Stream & MAE \\
CALCE~\cite{he2011calce} & Li-ion cell & Capacity & 12 & 3 & Stream & MAE \\
NASA CFRP~\cite{saxena2008cfrp} & Composite coupon & Stiffness & 3 & 3 & Stream & MAE \\
\midrule
\multicolumn{7}{l}{\emph{Anomaly detection and failure prediction}}\\
MIMII DUE~\cite{tanabe2021mimiidue} & Machine sound & Anomaly detection & 45 & 1 & Batch & ROC-AUC \\
CARE to Compare~\cite{gueck2024care} & Wind turbine & Anomaly screening & 36 & 3 & Batch & ROC-AUC \\
Backblaze~\cite{backblaze2024drive} & Hard disk & 7-day failure warning & 6$^{d}$ & 3 & Stream & Average precision \\
\bottomrule
\multicolumn{7}{l}{$^{a}$Records or operating conditions; the dataset has no independent units. \quad $^{b}$Simulated. \quad $^{c}$Batches of distinct vehicles. \quad $^{d}$Pools of drives.}
\end{tabular}
\end{table*}

\subsection{Protocol}
\begin{figure}[t]
\centering
\includegraphics[width=\columnwidth]{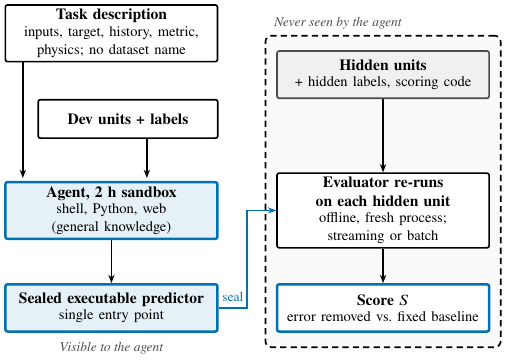}
\caption{Deployment-style evaluation. The agent sees the task description and the development units, works for two hours in a sandbox, and delivers a sealed, executable predictor. The evaluator runs it offline on each hidden unit; the hidden units, their labels, the scoring code, and the score itself remain invisible to the agent.}
\label{fig:protocol}
\end{figure}

An agent works as an engineer would on a new assignment. It receives a task description, which states the inputs, the target, the history that may be used at prediction time, the metric, and whatever physical information is available, but not the name of the dataset. It also receives the development units with their labels, a shell, and a Python environment. It then has two hours of wall-clock time, its own reasoning included, and may use the internet as an engineer would, but not to obtain evaluation data, labels, or existing answers. During development it may submit its predictor to the evaluator for a check that it runs, which reports errors such as an empty output but nothing about the hidden units. At the end it must deliver an executable predictor with a single entry point. Once the deliverable is sealed, an evaluator runs it offline on each hidden unit in a fresh process (Fig.~\ref{fig:protocol}), either in streaming mode, where each prediction may use only the records received so far, or in batch mode. The agent never sees the hidden labels or the scoring code, and nothing it reports about itself is taken on trust. A deliverable that does not run counts as a failure.

\subsection{Scoring}
Errors on different datasets are not comparable, so every fold is scored against a fixed baseline: the average loss of a random forest on generic time- and frequency-domain features and of a small one-dimensional convolutional network on the raw inputs, neither tuned to the task. With $E$ the loss of a method and $B$ that of the baseline (one minus the metric for accuracy, balanced accuracy, ROC-AUC, and average precision), the score is
\begin{equation}
S = 100\,\max\!\left(0,\; 1 - E/B\right),
\end{equation}
the share of the baseline's error that the method removes: $S=50$ means the error has been halved, and a failure scores zero. $S$ is best read comparatively: on a few folds the fixed baseline does worse than predicting a constant, which itself scores 3.9. Scores are averaged over the folds of a dataset and then over the 24 datasets with equal weight. Because $S$ does not register how much worse than the baseline a method can be, we also report how often a method fails or delivers a predictor worse than the baseline. On N-CMAPSS, the evaluator first omitted an input file that the task promised; all delivered predictors were rerun through the same public interface once it was fixed.

\subsection{Agents and References}
We evaluate four frontier agents, models too large to run on a plant's own hardware, GPT-6 Astra, Kimi K3, DeepSeek V4.1-Flash, and GPT-6 Luna, each run through the Codex command-line agent at its highest reasoning setting, and two open-weight models small enough for local hardware, Qwen3.6-35B-A3B and Gemma~4, run through OpenCode. All runs took place in September and October 2026, through Codex 0.156.1 and OpenCode 1.19.0 (Appendix~\ref{app:models}); on average, a run of the strongest agent took 32 minutes of active time and about 2.8 million tokens. As references we use the fixed baseline, two AutoML systems with and without automatic feature extraction, and faithful implementations of published methods.

Two independent audits checked the pipeline, one tracing each result from the paper back to the code and the other following the data from the raw files to the scores. Both reproduced the main scores to within 0.01 and the baseline scores to within a relative $2\times10^{-6}$, confirmed that development and hidden units do not overlap, and found no leak that changes a conclusion; the shortcuts they found are reported in Section~\ref{sec:shortcuts}.

\section{Agents Versus Published Methods and AutoML}
\label{sec:agents}
On this benchmark, the agents come out ahead of every reference we tried.

\subsection{Ranking and Stability}
\begin{figure}[t]
\centering
\includegraphics[width=\columnwidth]{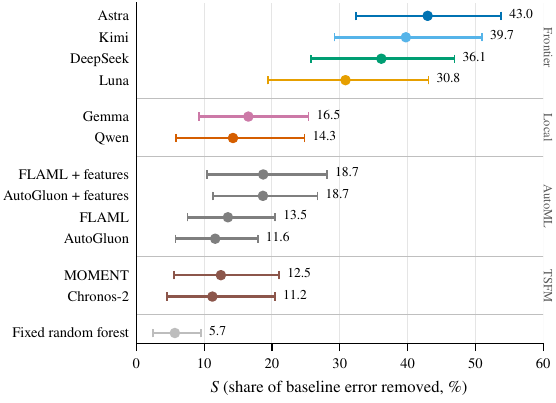}
\caption{Score $S$ over the 24 datasets, with 95\% intervals from a bootstrap over datasets. Frontier agents remove a third to nearly half of the baseline's error; local open-weight agents, AutoML, even with generic features, and time-series foundation models (TSFM) fine-tuned on each fold remain below 20. AutoML was not run on MIMII DUE and scores zero there; the fixed random forest is one of the two models that make up the baseline.}
\label{fig:scores}
\end{figure}

The four frontier agents remove between a third and nearly half of the baseline's error: $S$ ranges from 30.8 for Luna to 43.0 for Astra (Fig.~\ref{fig:scores}). The ranking is stable. Complete re-runs under identical conditions change no agent's average by more than 1.2 points and leave the order unchanged, and re-runs of five of the plan conditions of Section~\ref{sec:fit} move none by more than 2.2. The order also survives the obvious objections to our score (Table~\ref{tab:robust}): it is the same without the three datasets whose collection or preparation left a shortcut (Section~\ref{sec:shortcuts}) and on the 21 datasets with independent units, and when scores may fall below zero, only Astra and Kimi swap places, within each other's noise. Nor does it hinge on the fixed baseline: under eleven other choices, from either of its two models alone to the strongest AutoML pipeline, the four frontier agents keep their order in all but one, and the main comparisons of this paper keep their sign (Appendix~\ref{app:baselines}). Individual datasets are much noisier: two runs of the same agent can differ by more than 30 points on a single dataset, a point we return to in Section~\ref{sec:local}. Nor is the strongest agent infallible; nearly one run in ten delivers a predictor worse than the baseline.

Memory of public datasets does not explain the ranking either, although the agents do recognize them: asked to name the dataset from the task documentation alone, Astra identified 22 of the 24, and Luna and DeepSeek 18 each. What recognition offers, however, is the literature, and the literature is not what wins: faithful implementations of published methods lose to the agents' own designs (Section~IV-B). We class a dataset as well known if it is a standard benchmark with many published results; the other eight, classed after the runs, are less known (Appendix~\ref{app:models}). Against AutoML with generic features, the strongest agent gains about as much on the five less-known datasets whose collection or preparation left no shortcut (CFRP, MATWI, the NASA randomized battery, PHM Europe 2020, and SCANIA) as on the well-known ones (24 and 23 points), and the other frontier agents keep a lead there too, mostly a smaller one. The two agents whose reasoning we can read named the source dataset in about two of three runs, yet those runs did not score detectably higher, relative to the other agents on the same fold, than runs that did not name it.

\begin{table}[t]
\caption{Agent scores under alternative choices: over all 24 datasets; with scores allowed down to $-100$; without the three datasets whose collection or preparation left a shortcut (AHU, Petrobras 3W, and CARE; Section~\ref{sec:shortcuts}); on the 21 datasets with independent units; and the shares of runs that failed or delivered a predictor worse than the baseline, over the 58 runs of each agent's main experiment. In the floor column, failures still score zero. Ranks of the frontier agents in parentheses; 95\% intervals from a bootstrap over datasets.}
\label{tab:robust}
\centering
\footnotesize
\resizebox{\columnwidth}{!}{%
\begin{tabular}{@{}lcccccc@{}}
\toprule
Agent & $S$ & $S$, floor & $S$, without & $S$, 21 & Fail & Worse \\
 & (24) & $-100$ & shortcut sets & unit sets & (\%) & (\%) \\
\midrule
Astra & 43.0 (1) & 33.8 (2) & 40.4 (1) & 37.9 (1) & 0.0 & 8.6 \\
 & {\scriptsize[32.4, 53.8]} & {\scriptsize[15.5, 50.0]} & {\scriptsize[29.4, 52.0]} & {\scriptsize[28.3, 46.8]} &  &  \\
Kimi & 39.7 (2) & 37.7 (1) & 38.4 (2) & 34.7 (2) & 1.7 & 8.6 \\
 & {\scriptsize[29.2, 51.0]} & {\scriptsize[26.6, 49.5]} & {\scriptsize[26.6, 50.7]} & {\scriptsize[25.1, 44.1]} &  &  \\
DeepSeek & 36.1 (3) & 30.9 (3) & 34.7 (3) & 31.0 (3) & 0.0 & 10.3 \\
 & {\scriptsize[25.7, 46.9]} & {\scriptsize[18.7, 43.5]} & {\scriptsize[24.2, 46.0]} & {\scriptsize[22.6, 39.7]} &  &  \\
Luna & 30.8 (4) & 17.5 (4) & 29.2 (4) & 24.4 (4) & 1.7 & 20.7 \\
 & {\scriptsize[19.4, 43.1]} & {\scriptsize[$-$2.3, 35.7]} & {\scriptsize[17.5, 41.9]} & {\scriptsize[15.0, 34.7]} &  &  \\
\midrule
Qwen3.6 & 14.3 & 0.8 & 14.4 & 8.3 & 29.3 & 27.6 \\
 & {\scriptsize[5.9, 24.8]} & {\scriptsize[$-$13.5, 15.8]} & {\scriptsize[5.4, 25.7]} & {\scriptsize[3.6, 13.6]} &  &  \\
Gemma 4 & 16.5 & 0.0 & 16.7 & 12.4 & 3.4 & 32.8 \\
 & {\scriptsize[9.2, 25.4]} & {\scriptsize[$-$15.6, 14.6]} & {\scriptsize[9.3, 25.5]} & {\scriptsize[7.1, 18.1]} &  &  \\
\bottomrule
\end{tabular}}
\end{table}

\subsection{Published Accuracies Do Not Travel}
A natural reference for an agent is the literature itself. We selected 20 papers on four benchmark datasets, C-MAPSS, N-CMAPSS, Paderborn, and Tennessee Eastman, by a fixed rule: on each dataset, the most cited method papers since 2019 that name it and train on labeled data, plus the most cited since 2024, with three N-CMAPSS papers taken from a recent reproduction collection~\cite{theiler2026paper}; Appendix~\ref{app:lit} lists them. Eighteen of the 20 are IEEE journal papers, because copies from other publishers could not be retrieved automatically, and three of the five Paderborn papers come from one research group. The strongest agent read each paper and implemented its method faithfully on our folds, within the same two hours, using the paper's own results only to understand the method. An independent LLM judge, Claude Opus 5.5, rated all 30 implementations as faithful. Yet only 6 of the 30 implementations, from 6 of the 20 papers, beat the agent's own design for the same fold. On average, the published methods fall short on C-MAPSS, N-CMAPSS, and Paderborn and draw level on Tennessee Eastman; the best of them, however, beat the agent's design on Tennessee Eastman and N-CMAPSS (Appendix~\ref{app:lit}).

\begin{figure}[t]
\centering
\includegraphics[width=\columnwidth]{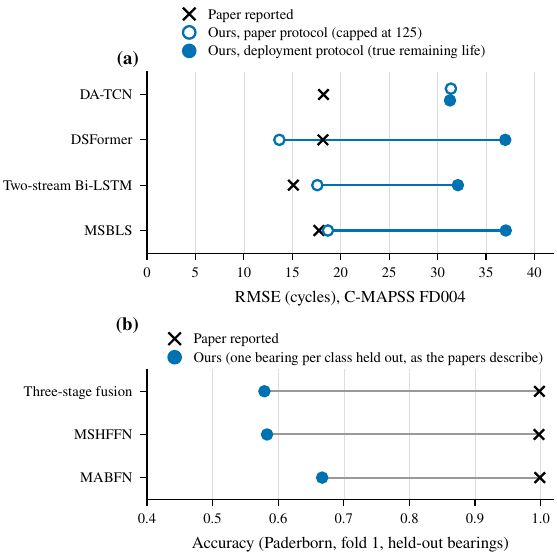}
\caption{Faithful implementations of published methods, under the papers' own protocols and under ours. (a) C-MAPSS FD004, with DA-TCN~\cite{song2021datcn}, DSFormer~\cite{li2024dsformer}, a two-stream Bi-LSTM~\cite{jin2022bilstm2stream}, and MSBLS~\cite{xu2024msbls}: implemented under the papers' protocol, which caps the remaining life of test engines at 125 cycles, three of the four come close to their published errors; implemented and scored on the true remaining life, as deployment requires, the same methods err about twice as much. For DA-TCN the two protocols give nearly the same error, so its two markers are drawn slightly apart. The fifth C-MAPSS method of Appendix~\ref{app:lit}, a dual-task LSTM, was not reproduced under the papers' protocol and is omitted. (b) Paderborn, with three fusion networks~\cite{wang2021tim3stage,wang2025mshffn,wang2023mabfn}: with one bearing per class held out, as the papers describe, the implementations reach none of the reported accuracies.}
\label{fig:protocols}
\end{figure}

Were the implementations simply poor? Under the papers' own protocols, and with four hours instead of two, the same agent reproduces the reported numbers in eight of fourteen attempts: six of eight on C-MAPSS (four methods on two subsets), one of two on Tennessee Eastman, the single attempt on N-CMAPSS, and none of three on Paderborn. A single, ordinary convention can make the difference. C-MAPSS papers, like our benchmark, score each test engine at its last observed cycle, but they cap its remaining life at 125 cycles before computing the error. Under that convention, three of the four implementations on our engines come close to the published errors, at 13.7 to 18.7 cycles; implemented for the true remaining life, as deployment requires, the same four methods err by 31 to 37 cycles on the very same engines (Fig.~\ref{fig:protocols}). The gap is not only early-life error: on engines within 125 cycles of failure, where capping changes nothing, the implementations still err by 25 to 34 cycles. On Paderborn the gap is starker. Holding out one whole bearing per class, as we read the papers' descriptions, methods reported at 99.7--99.9\% accuracy reach 58--67\%, and every one of them mistakes the held-out inner-race bearing for an outer-race fault; we could not obtain the authors' own splits to check this reading. Published accuracy tells us how well a method fits a protocol; it does not tell us how well it fits the next machine.

\subsection{AutoML and Foundation Models Are Not Enough}
A second objection is that any automated search would do as well. It does not. Given the same inputs as the fixed baseline and a two-hour search target, AutoGluon and FLAML reach $S$ of 11.6 and 13.5, closer to the local open-weight agents than to any frontier one. Their inputs, the baseline's own, subsample long recordings and cap the number of training rows, and a failed fit scores zero, as it does for the agents; scoring a failed fit as a constant prediction instead raises both by one to two points. A bank of generic signal features lifts both to 18.7, and tsfresh features~\cite{christ2018tsfresh} lift FLAML to 28 on seven signal datasets, where Astra scores 57 on the same folds. AutoML also tends to overrun: nearly every AutoGluon fit took longer than two hours. Overruns were allowed and scored; only a fit stopped by the cluster's job limit counted as a failure. Generic features close the gap where the problem is easy, taking CWRU from 0 to above 90, but not where it is hard: on the hydraulic rig, the IMS bearings, Paderborn, and SEU, AutoML stays below 15 while Astra scores between 21 and 100. What the agents add is not search but judgment about the machine at hand.

Nor does a general model. Time-series foundation models are the closest PHM has to the general models of language and vision. We fine-tuned two whose published pretraining corpora contain none of our datasets, MOMENT~\cite{goswami2024moment} and Chronos-2~\cite{ansari2025chronos2}, on every fold, with the same downsampled inputs as AutoML and one fixed pipeline, the way a general model is meant to be used; tuning them to each dataset might lift them, but that is a modeler's work. They reach $S$ of 12.5 and 11.2, level with AutoML without features and far below every frontier agent (Fig.~\ref{fig:scores}). They do well on some bearing datasets, CWRU and XJTU-SY among them, and score below 2 on ten or more others, including C-MAPSS and Tennessee Eastman. The agents seem to agree: a foundation-model library was installed in their environment, yet no agent in any of our runs imported one, although they did use pretrained audio and vision backbones. The next section asks where the agents' judgment lives.

\section{Where the Agent's Advantage Lives: Fit, Not Size}
\label{sec:fit}
Section~\ref{sec:agents} leaves a question open. When an agent's design beats published models on a machine it has just met, is the agent a better programmer, or did it make better choices for that machine? We answer by transplanting designs from one agent to another, by having agents design without the data, and by stripping designs of heavy machinery.

\subsection{Design}
After a run, the method an agent delivered on a fold is written up as a plan by a separate session of GPT-6 Astra, which reads the delivered code and the commands of the development run under fixed rules but never the data. A plan has seven short sections, from task framing and features to model, validation, and inference, and stays within 900 Chinese characters, about 500 English words. It contains no code and no number fitted to the data, but it states every choice the agent made, down to the hyperparameters it selected by trial on the fold's development data. A plan is, in effect, a model specification without its weights: much like the method section of a paper, except that it was written for one machine (Appendix~\ref{app:plan}). A second agent receives the plan, read-only, together with the usual task and is asked to implement it, deviating only where it conflicts with the task; it is not told whose plan it is. We cross the strongest and the weakest of the four frontier agents, GPT-6 Astra and GPT-6 Luna, in both directions, so that plans move across the widest gap; these two carry the later experiments. As a larger and a smaller model from the same developer, the two likely share much of their training and style and differ mainly in capability, which isolates capability for the comparison; Section~\ref{sec:local} hands the same plans to a local model. Plan runs use the OpenCode harness, and their controls are runs without a plan in the same harness, which score within 1.2 points of the main runs through Codex (44.2 against 43.0 for Astra, 31.0 against 30.8 for Luna).

\subsection{Fit Travels, Good and Bad}
\begin{figure}[t]
\centering
\includegraphics[width=\columnwidth]{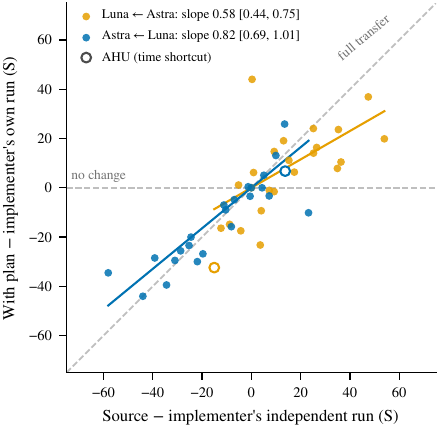}
\caption{Fit travels, good and bad. Each point is a dataset: how much better the plan's author scored than an independent run of the implementer (horizontal), against how much the implementer's score moved, relative to its own run, when it followed the plan (vertical). Both transplants move toward the plan's author, upward or downward. Slopes are fitted through the origin (Luna following Astra, 0.58; Astra following Luna, 0.82) with 90\% bootstrap intervals; noise in the independent runs biases them toward zero. Hollow markers show the AHU dataset, whose time shortcut travels with the plans.}
\label{fig:transplant}
\end{figure}

Given Astra's plans, Luna moves toward Astra (Fig.~\ref{fig:transplant}). On average its score rises from 31.0 to 36.8, closing nearly half of its gap to Astra, and the pattern across datasets is unmistakable: on the 18 datasets where Astra was the better modeler, Luna gains 12.2 points; on the 6 where Astra was worse, Luna loses 13.3, following Astra down.

The reverse transplant is starker. Given Luna's plans, Astra falls from 44.2 to 31.5, about half a point from Luna's own score; it drops on 17 of 24 datasets, by as much as 44 points, and where Luna had been the better modeler, Astra rises with it. On 16 of the 44 folds where the two agents' scores are not both at the floor or the ceiling, Astra's predictions are identical to Luna's, and on 3 more nearly so, although it wrote every line of its code anew, shares only 1--3\% of its lines with Luna's code, and could not see Luna's run. It also finished in a median of 12 minutes, against 42 for its own runs in the same harness. Run a second time, each transplant scored within a point of the first (37.8 and 31.5).

What is striking is not that a specification can be reimplemented, but how little it takes. A plain-text plan of a few hundred words, written for one machine, carries most of the gap between two frontier agents, and once the design is fixed, a stronger implementer adds almost nothing to it, while a weaker one realizes much of it. A published method is such a specification too; these plans differ only in having been chosen on the machine they are applied to. Following a plan does not by itself cost an implementer anything: following its own plans, even lightened, Astra scored 42.6 against 44.2 without a plan (Section~\ref{sec:lighten}). Astra fell with Luna's plans because of what they said; and the half of the gap that Luna left open with Astra's plans lies in carrying them out, not in what the plans lost (Section~\ref{sec:local}).

\subsection{Close Before the Data}
\label{sec:blind}

Could the strongest agent have written those few hundred words without the data? We asked each frontier agent to write a plan for every fold from the task description, the machine's documentation, and a card listing the layout of the data (its files, fields, shapes, and sampling rates) but not a single value. A fresh session wrote each plan under the same template, with no tools and no access to files; both agents then carried out every plan on the data, under the same instructions as before.

Before touching the data, it hardly mattered whose plan an agent carried out (Table~\ref{tab:blind}). Holding the implementer fixed, Astra scored 29.5 with its own plans written without the data and 28.0 with Luna's, and Luna 27.6 with Astra's and 26.0 with its own: about 1.5 points either way. Second runs give 1.4 for Astra and, against a rerun of Luna's own plans, 3.8 for Luna. Plans written on the data were another matter: to the same implementer, Astra's were worth 5.8 points more than Luna's when Luna carried them out and 12.7 points more when Astra did. Designing before seeing the data cost Astra 14.7 points and Luna 5.0; given the same data, the stronger agent made far more of it. Given to Luna as advice rather than as instructions, Astra's plans changed little. A plan written on the data pins the model down: following Luna's original plans, Astra reproduced Luna's predictions on 16 of 44 folds. A plan written without the data leaves more to its implementer: two agents carrying out the same such plan made identical predictions on only 2 to 5 of 41 to 44 folds, although their scores differed by just 2 points.

Knowledge carried the plans furthest where the field knows the machine. On the 16 best-known datasets, the plans written without the data came within 8 to 12 points of Astra's own designs; on the eight less-known ones, they fell 27 to 30 points short (15 to 23 without the shortcut sets; Section~\ref{sec:shortcuts}). Knowledge without the data, at about 28, search without judgment, at 12 to 19 for AutoML, and judgment applied to the data, at about 44 for the strongest agent, are three different things. What sets the strongest agent apart is not what it knows but how much it learns from the data in front of it.

\begin{table}[t]
\centering
\caption{Score by who wrote the plan, whether its writer had seen the data, and who carried it out ($S$ over 24 datasets).}
\label{tab:blind}
\begin{tabular}{@{}llcc@{}}
\toprule
 & & \multicolumn{2}{c}{Carried out by} \\
\cmidrule(l){3-4}
Plan by & Data seen & Luna & Astra \\
\midrule
Astra & yes & 36.8 (37.8) & 44.2$^{a}$ \\
Luna & yes & 31.0$^{a}$ & 31.5 (31.5) \\
Astra & no & 27.6 & 29.5 (30.0) \\
Luna & no & 26.0 (23.8) & 28.0 (28.6) \\
\bottomrule
\multicolumn{4}{@{}l}{$^{a}$The agent's own run, without a plan.}\\
\multicolumn{4}{@{}p{0.95\columnwidth}@{}}{In parentheses, a second run of the same condition; unless stated otherwise, the text uses the first run.}
\end{tabular}
\end{table}

\subsection{Size Is Not the Source}
\label{sec:lighten}
We then asked Astra to lighten its own plans under written rules: keep the task framing, features, validation, and selected hyperparameters; replace packages that are not preinstalled with scikit-learn; merge ensembles into a single model; drop neural networks; and keep each training run under ten minutes on four CPUs. Following these lightened plans, Astra scored 42.6, against 44.2 without any plan. Only 6 of its 58 original designs had used a neural network at all; most combined several models, and where a design leaned on such machinery, lightening it cost some points. But most of the value survived. What carries the score is the fit of a design to its machine, not the size of its model.

\subsection{Only the Design Moves the Score}
\label{sec:levers}

\begin{figure}[t]
\centering
\includegraphics[width=\columnwidth]{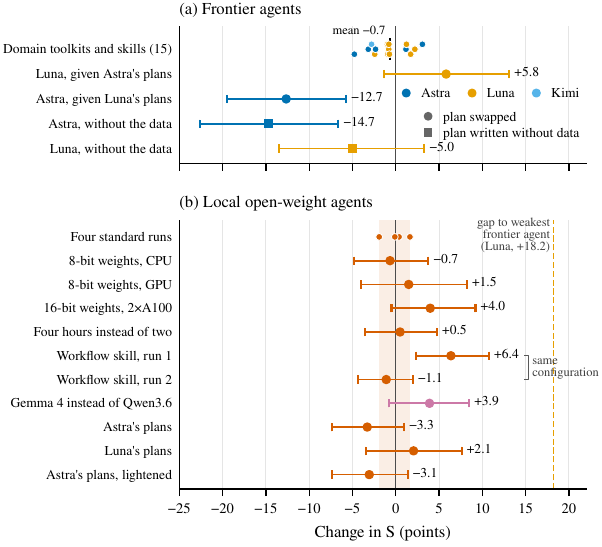}
\caption{What moves the score. (a) Seven toolkits, tried in fifteen agent--toolkit combinations, leave the frontier agents' scores where they were; swapping the plan moves them by as much as 12.7 points, up or down (circles), and carrying out a plan written without the data costs the strongest agent 14.7 (squares). (b) For local open-weight agents, nothing closes the 18-point gap to the weakest frontier agent (dashed), and two runs of the same configuration differ by 7.5 points; the shaded band spans four standard runs of Qwen3.6. Changes are measured against the agent's own run without the intervention in (a) and against the mean of the four standard runs in (b). Intervals are 95\% from a paired bootstrap over datasets and do not include run-to-run variation.}
\label{fig:levers}
\end{figure}

If the advantage lies in fit, capability added from outside should not move it, and it does not. Over the project, we gave the frontier agents seven toolkits, tried in fifteen agent--toolkit combinations (Appendix~\ref{app:toolkits}), each on the datasets it applies to: a public predictive-maintenance skill from Amazon Web Services~\cite{awspdmskill}, an open-source predictive-maintenance server with diagnostic workflows~\cite{dimaggio2026pdmmcp}, tsfresh, the anomaly-detection library PyOD~\cite{zhao2019pyod}, NASA's prognostics packages ProgPy~\cite{progpy}, and feature components and validation tools of our own. No toolkit helped every agent it was given to. On average, they changed the score by $-0.7$ points, and the only clear change was a loss: Astra kept tsfresh features in three of four runs and scored 4.8 points lower (Fig.~\ref{fig:levers}a). Two of the toolkits never made it into a final model at all. Against all of this, the plan transplants stand apart: a few hundred words moved the same agents by $+5.8$ and $-12.7$ points. Designing without the data moved Astra most of all, by $-14.7$ points, and Luna by $-5.0$ (Section~\ref{sec:blind}).

\section{Only to Implementers That Can Execute Them}
\label{sec:local}
Models small enough to run inside a plant are where PHM agents would be most useful, and they are far behind. Qwen3.6, running on a single workstation, scores 14.3 in its main run, against 30.8 for the weakest frontier agent and on a par with AutoML (Fig.~\ref{fig:scores}); it fails on 17 of 58 runs, almost always by running out of time. Gemma~4 fails differently: it delivers on 56 of 58 runs, but what it delivers is weak, and its score of 16.5 is barely higher.

\subsection{Plans Make Them Deliver Less}
Handing them the strongest agent's design does not help. Given Astra's plans, Qwen3.6 delivers a working predictor on only 32\% of runs, against 57\% without a plan (Fig.~\ref{fig:execution}a), and its score falls from 12.6, the mean of its four runs without a plan, to 9.3; two in three of its runs end in a timeout, lost in code, dependencies, and repair loops. Luna's simpler plans hurt far less: Qwen delivers on 53\% of runs and scores 14.7 (Fig.~\ref{fig:execution}a). What matters is the load of a plan, not its author. We index load by five features of the author's own implementation: the number of models, external dependencies, the use of a neural network, compute-heavy time, and lines of code. On light and middle plans, Qwen delivers as often on Astra's as on Luna's, and on light plans at least as often as with no plan at all; but Astra's plans pile up in the heaviest third, 64 of 116 against 12 of Luna's, and there Qwen delivers on only one run in five. Nor does lightening rescue it. Given the lightened plans of Section~\ref{sec:lighten}, which it read in almost every run, Qwen still delivered on only 31\% of runs and scored 9.6. The trouble is not hardware either: the drop persists on folds whose source run used no GPU.

\subsection{Capability Buys the Probability of Completion, Not Its Quality}
\begin{figure}[t]
\centering
\includegraphics[width=\columnwidth]{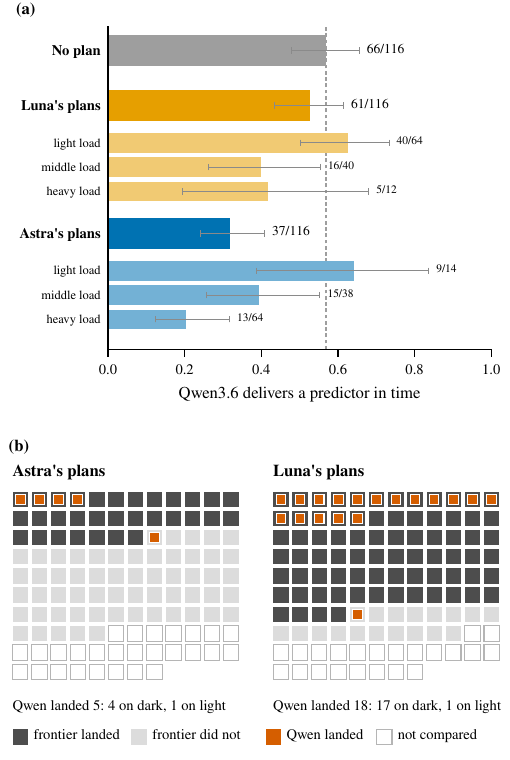}
\caption{Plans reach only implementers that can execute them. (a) Share of Qwen3.6's runs that deliver a working predictor within the two hours, without a plan (two runs) and with Luna's or Astra's plans (two runs each; thick bars). Thin bars split each plan source by thirds of plan load, cut over the plans of both sources; the dashed line marks delivery without a plan, and whiskers are 95\% Wilson intervals. (b) Each square is one Qwen run on one fold, 116 per plan source. An implementer lands a plan when it delivers and reaches the plan author's own result; a run that does not deliver does not land. Dark squares mark folds where the frontier implementer (Luna for Astra's plans, Astra for Luna's) landed the plan, light squares folds where it did not, and orange marks Qwen's landings: 21 of its 23 fall on dark squares. Hollow squares are not compared, because both implementers scored 100, both scored 0, or the plan's author scored 0, so landing carries no information.}
\label{fig:execution}
\end{figure}

Yet what Qwen does deliver is no worse than what it builds on its own: on folds where it delivers both with and without Astra's plans, its scores differ by about one point, and the lightened and original plans give the same quality. What separates it from the frontier agents is whether it completes a design, that is, delivers a working implementation within the two hours, not how well. Success is also nested: Qwen lands a plan, delivering it and reaching its author's result, almost only on folds where a frontier implementer lands it too (Fig.~\ref{fig:execution}b). Capability, in other words, buys the probability of completing a design, not the quality of what is completed.

\subsection{Nothing Else Closes the Gap}
Nothing we tried closed the gap of about 18 points between Qwen3.6's standard runs and the weakest frontier agent (Fig.~\ref{fig:levers}b). Higher-precision weights, from 4-bit to 8-bit to full 16-bit, moved the score by a few points in directions that did not repeat. Giving the agent a GPU changed little, because it barely used one. Doubling the time to four hours cut failures from 23 to 12 of 58 runs, against a run made at the same time, but left the score where it was. A workflow skill that guides the agent step by step helped by 6.4 points in one run and not at all in its repeat. That repeat is the most instructive number in this section: the same configuration, run twice, differed by 7.5 points, more than any lever we pulled. Comparisons of single runs of local agents, ours included, should be read with this in mind.

Reading its failures shows what is missing. Of 100 failed runs we read in full, half were slow but making progress: a single training or feature command consumed most of the budget, or the agent tuned methods before securing a version that worked. A third were stuck in loops, nearly all on the same complaint from the evaluator, an empty output, because the agent never worked out how its predictor would be called. Repetitive text caused none, and 45\% of the runs that timed out never once submitted a deliverable for checking. What a local model lacks, then, is not knowledge of what to build: handed the best design, even one stripped to standard tools, it delivers less. It lacks the engineering judgment to build that design within the time it has: estimating what a step will cost, securing a working version first, and learning the interface it must serve. These findings hold at our fixed budget of two hours, which sets where the threshold lies.

This is also why local modelers are worth improving even if they never match frontier ones. In plants whose data may not leave the site, engineers already work the way our plans written without the data were made: they bring knowledge in from outside and apply it inside. On its own, such knowledge was worth no more than what a weaker frontier agent achieved with the data, and given as advice rather than as instructions, it changed little (Section~\ref{sec:blind}). The rest of the value is made on the data, and only a modeler beside the data can make it. A local model need not match a frontier one to be worth having; what it needs is not more knowledge but the capacity to act on the data in front of it.

\section{Discussion}

\subsection{What Counts as a Contribution in the Agent Era}
The worry that researchers overfit their designs to the nominal test sets of benchmarks is decades old~\cite{dwork2015reusable,recht2019imagenet,lipton2018troubling}. In the agent era it becomes acute. Picture a researcher who wants better numbers on a few datasets. With an agent at hand, trying one more popular module, and then another, costs minutes rather than weeks; sooner or later some combination improves the results on the researcher's own test split, and yet another XXX-Net is published as a general model. Such a model is overfitted, not to its training data but to a protocol, and every one of them makes it harder for the field to see which ideas carry over to the next machine. Our results show how thin the line can be: a single, ordinary convention, capping the remaining life of test engines, can separate published C-MAPSS errors from errors roughly twice as large on the very same engines (Section~IV-B).

To treat an agent as a researcher is to make research itself the object of study. An agent can be run again, denied the data, handed another agent's plan, or given a toolkit, and every choice it makes leaves a trace; no such experiment can be run on the researchers whose methods fill our journals. What the agents did, we therefore read as a magnified picture of the field. They recognized the famous datasets and knew their literature, yet the published methods did not travel; they fitted whatever they were given, the way the data were collected included; and their own validation could confirm only what their data allowed them to see. These are not the agents' failings alone. They are the field's, made visible and measurable.

Agents have greatly multiplied what a researcher can do. Yet researchers, like the agents themselves, have little sense of how good the things they make really are. This is not a personal failing but a structural one: a researcher's own split, like an agent's own validation, can only confirm what its data allow it to see. It is also why benchmarks exist; they supply the self-knowledge that neither researchers nor agents possess on their own.

In PHM, the problem is compounded by a scarcity of data that is easy to overlook. Industrial data appears abundant, yet the physical units behind it are few. CWRU, still among the most widely used datasets, records artificially seeded damage on a handful of bearings, and its leaderboards passed 99\% long ago~\cite{hendriks2022cwru}. Unlike images or text, which humans perceive on a common scale, a PHM signal means nothing without its sampling rate, its sensor, and its machine; two waveforms of the same shape can mean entirely different things. Each dataset therefore covers a narrow distribution, and no single dataset can tell how a model will fare in the field. This is also one reason why the apparent ocean of industrial data has not yet yielded a foundation model that does for PHM what large language models did for text. What the field needs is a common benchmark that spans many distributions: shared, so that differences in evaluation no longer masquerade as differences in methods, and diverse, so that generality is actually put to the test. Such a benchmark can test the generality of a model or, as in this paper, of a modeler.

What if the domain knowledge built into frontier models already rivals, or exceeds, that of most human experts? We did not set out to measure this, but our results point that way. Before touching the data, the strongest and a weaker frontier agent designed within a few points of each other (Section~\ref{sec:blind}), as if both stood on nearly the same plateau of domain knowledge; domain toolkits and skills written by people added nothing to either on average (Fig.~\ref{fig:levers}a); and the agents recognized most datasets from their documentation alone. Much current work measures the PHM knowledge of general models or adds to it~\cite{phmbench2025,diagent2026}. For frontier models, that knowledge may already be in place, and further skills may no longer help. The plateau has an edge, however: knowledge carried furthest on well-known machines and much less far on unfamiliar ones. What is scarce is not knowledge but what the strongest agent did better than the rest, learning from the data of one machine, and the means to tell whether what it learned was right.

Our results echo Sutton's bitter lesson~\cite{sutton2019bitter}: in the long run, methods that leverage computation beat methods built on human knowledge. Knowledge written in by people did not move the agents. Toolkits left their scores where they were, published methods lost to their designs, and plans written from knowledge alone, without the data, scored about 28. What set the strongest agent apart was what it found in the data. AutoML is no exception to the lesson: it searches, but over a menu that people wrote. At the level of the model, the opposite lesson holds. The no-free-lunch theorems state that, averaged over all problems, no optimizer beats another, so any advantage comes from matching the structure of a problem~\cite{wolpert1997nfl}, and designs fitted to each machine did beat the general ones. The agent is a general method whose products are specialized ones. Moving generality from the model to the modeler lets both lessons hold at once.

For researchers, this changes what is worth contributing. When an agent can produce a new ``what'', a new architecture or a new model, in an afternoon, the contributions that move a field forward are the ``why'' and the ``how'': why a method works on one machine and fails on the next, and how to tell the difference before it is deployed. A gain on one's own split entertains its author; an explanation that holds on the next machine moves the field. Building a model for each machine is not the same as overfitting it: overfitting tunes a model to the very data on which it is judged, whereas the designs here were chosen on development units and judged on units no one had seen.

\subsection{Capability as Strength, Design as Stress}
\label{sec:lens}

Our observations fit a lens familiar to reliability engineers. In stress--strength interference, a component fails when the stress it meets exceeds its strength~\cite{oconnor2012reliability}; in item response theory, a person answers an item correctly when ability exceeds difficulty~\cite{rasch1960}. Read an implementer's capability $\theta$ as strength and the demand $c$ of a design as stress, and the score the implementer can expect from that design becomes
\begin{equation}
\bar S = \Phi\!\left(\frac{\theta - c}{\sigma}\right) Q,
\end{equation}
the chance that the design is completed, with $\Phi$ the standard normal distribution function and $\sigma$ the spread of both, times $Q$, what the completed design is worth. In this light, three of our findings fall into place. Capability enters only the first factor: it buys the chance of completing a design, while the worth of a completed design belongs to the design. Completions are nested, because a design that stops a strong implementer stops a weaker one too. Re-runs show the shape of $\Phi$ directly: following Luna's plans, far within its strength, Astra landed or missed the same folds in both runs on 56 of 58; following Astra's heavier plans, near its strength, Luna did so on only 47, and 11 folds switched between runs. Far from the threshold the outcome is settled; near it, it is close to a coin toss. And lightening a design helps only if it lowers the stress that matters: stripping heavy machinery left most of the design's demand in place, and the local model failed as before. Our plans written without the data point to a second kind of strength. What a designer brought before touching the data separated the frontier agents far less; what it learned from the data separated them most, and that strength enters $Q$, the worth of a design, not the chance of completing it. The lens also has a term we have not measured. An agent chooses its design from an estimate of its own strength, $\hat\theta = \theta + \delta$; when $\delta$ is large, it picks designs it cannot complete, or fails to recognize the best of those it can. That $\delta$ is the self-knowledge discussed above, and measuring it is our next step.

\subsection{Fitting the Machine, or the Dataset}
\label{sec:shortcuts}
One dataset revealed something the score alone could not. In the AHU dataset, faults were introduced in all air-handling units on the same schedule, so the timestamp alone predicts the fault. Most frontier agents found this regularity and used it, and it travelled from one agent to another inside their plans. Using it is not cheating: the task allowed it, and no general model could have known it. It is customization at full strength. But the regularity belongs to how the data were collected, not to the machine. Held-out units cannot expose it, because they share the schedule; a real building would expose it on the first day. When we ran the sealed predictors again with each unit's clock shifted by a random offset, scores of up to 86 fell to about 18 or less, and the gap between frontier and local agents on this dataset all but vanished. Nor must the shortcut be sought: plans written without the data that added the hour of the day and the day of the week, as building engineers routinely do, scored up to 37 on this dataset, against at most 16 for plans without them.
Two less-known datasets carried subtler traces of how they were made. In Petrobras 3W, our benchmark requested predictions only for the labeled rows of each instance, so where the first request falls within an instance, together with the instance's length, identifies one class; in CARE, the length of the observed window separates anomalies within a wind farm. Astra's designs turned both into features. Agents, like researchers, find and use whatever traces the making of a dataset leaves behind, and these traces belong to the benchmark, not to the machines. Without the three datasets, the same implementer gains about 2.5 points from Astra's plans written without the data and 6.1 to 11.6 points from its plans written on the data, against 1.5 and 5.8 to 12.7 with them; on the less-known datasets, plans written without the data fall 15 to 23 points short of Astra's own designs rather than 27 to 30. No ranking or comparison changes direction.

Generality forces a model to rely on what holds everywhere; locality lets a modeler use whatever holds where it works. That freedom is the point of customization, but a correlation can hold here because of the machine's environment, as when a building's schedule drives its load, or because of how the data were made, as with the shared fault schedule above. A customized model fits whatever it is given, the machine and the making of its data alike, and only knowledge from outside the data can tell the two apart. We keep these datasets in the benchmark as hidden indicators of that limit: a high score on them, earned through such traces, signals a fit to how the data were made that a held-out test cannot see.

Our own documentation held one shortcut too. For Paderborn, a table of each bearing's operating history, keyed by the identifiers used at deployment, revealed which bearings were healthy; one of 81 delivered predictors used it, without gain, and removing Paderborn changes none of our comparisons.

\subsection{Limitations}
Our evidence has limits. Most configurations were run once; aggregate results are stable across complete re-runs, but single datasets are not, so per-dataset contrasts should be read as descriptive, and our intervals reflect variation across datasets, not across runs. The datasets are public and may have been seen during pretraining: the agents were not told which dataset they faced, but they recognize most of them from the task documentation, and the reasoning of two of them cannot be inspected, so memorization cannot be ruled out entirely. That the published methods were implemented faithfully rests on a single LLM judge. Agents could also use the internet during development, as a practitioner would: public documentation, papers, code, and pretrained models were allowed, while obtaining evaluation data, labels, or existing answers was forbidden but not technically blocked. Scanning the commands and web-tool calls of every development run reported here, we found one violation: Kimi looked up the duration of a hidden IMS test and hard-coded a wrong value, which lowered its score on that fold. A few searches for the benchmark's own interface and error messages returned nothing; all other network use installed packages, downloaded generic pretrained weights, or consulted documentation, code, and public descriptions of the datasets, and none retrieved evaluation data or labels. The scan sees only addresses written in commands or printed in their outputs. The agents ran in their usual harnesses, Codex for the frontier agents' main runs and OpenCode for the local agents and the plan runs, so what we compare are agents as deployed, not bare models. The local models in the main experiment ran on CPUs only, although runs with GPUs changed little. Finally, our plans are specifications written after the agent's own trial and error, within 900 characters; they transfer a design chosen on the data. The plans were written in Chinese, which all implementers handle, and all transplanted plans, Luna's included, were written up from the delivered code by a session of Astra; if anything, this should favor Astra as an implementer, yet given Luna's plans, Astra fell to Luna's level. The plans written without the data were run days to weeks after the runs they are compared with, and Luna wrote each of its plans in about a minute and a half where Astra took about ten; if anything, this should favor Astra's plans, yet Astra's plans led by only 3.5 to 6.2 points. Whether plans travel as well between frontier models from different developers we have not tested, and the closeness of the two agents before touching the data may partly reflect their shared origin. Above all, our setting assumes labeled data from other units of the same kind of machine. When a new type of machine arrives with no labels at all, the cold start that motivates much of transfer learning, there is nothing to fit, and whether a modeler can still help there remains open; our plans written without the data show only what knowledge contributes before any data is seen, more on well-known machines and less on unfamiliar ones.

\subsection{Future Work}
Three directions follow. The first is planning without data for local implementers. With frontier implementers, a plan written without the data was worth about what a weaker agent achieves on its own; whether such a plan helps a local model on site, in the setting that motivates it, is open, and so is how far newer local models or longer budgets move the threshold that stops them. The second is reuse across tasks: whether the decisions written for one machine help on another, which would turn a modeler's experience into something like a foundation and offer a route toward the cold start. The third is self-knowledge: whether an agent can tell a good plan from a bad one, or its best attempt from the rest, without a benchmark to tell it. A laboratory test rig whose data have never been published would offer a lockbox test for all three.

\section{Conclusion}
We asked whether, in the age of agents, generality should reside in the model or in the modeler. Evaluated the way models are deployed, general-purpose agents outperformed AutoML averaged over 24 PHM datasets and, for the strongest of them, faithful implementations of published methods on average on three of four datasets. Their advantage lay less in the size of their models than in fit: decisions made on each machine's data that fit in a plain-text plan of a few hundred words, decisions that travel to any implementer able to carry them out, but not yet to the models that fit on local hardware. Before touching the data, it mattered little whose plan an agent carried out; what set the strongest apart was how much it learned from the data in front of it. PHM, we believe, is moving from publishing models to producing decisions. The evolution of its methods has not reached an end; it has turned a corner. Its next step returns to building a model for every machine, only now the builder is a general modeler. One route to generality, in other words, is not to build a general model, but an agent that adapts methods to each setting. In studying agents as researchers, we found ourselves studying research itself.

\appendices
\section{An Example Plan}
\label{app:plan}
Plans were written in Chinese, within 900 characters; we translate one here. It describes the method Astra delivered on the second fold of PHM 2010, where the task is to track the flank wear of a milling cutter cut by cut. The plan states every decision and every hyperparameter Astra selected, but no fitted coefficient: the implementer must write the code and fit the models itself. Given this plan, Luna reproduced Astra's score exactly, and Qwen3.6 came within a tenth of a point in the second of its two runs; in its first run, it delivered a predictor worse than the baseline.

{\footnotesize
\noindent\textbf{Task framing.} Stream-wise regression per cutter; each cut is one sample; predict the flank wear averaged over the three flutes after the cut, in \textmu m; the history uses only cuts already received.

\noindent\textbf{Data and preprocessing.} Use the three vibration axes only. Take 20\%--80\% of the current record (bounds floored, keeping at least one point); subsample with step $\lfloor$points/20000$\rfloor$, minimum 1. Read in single precision, compute in double; set non-finite values to zero, clip to $\pm10^6$, and subtract each axis's mean. Raise an error on empty signals.

\noindent\textbf{Features.} (1) Difference amplitude: population standard deviation of $(X-Z)/\sqrt{2}$. (2) Projection amplitude: project onto the unit directions $(1,-1,0)/\sqrt{2}$ and $(1,1,-2)/\sqrt{6}$ and take the square root of the smallest eigenvalue of the two-dimensional population covariance. (3) Minor-axis amplitude: square root of the smallest eigenvalue of the unbiased three-axis covariance (zero for a single point); negatives are set to zero before the square root. Each model takes two inputs: the amplitude raised to an inner power, accumulated over the cutter's cuts and raised to an outer power; and a run-in term $1-\exp(-\text{cut}/\text{scale})$.

\noindent\textbf{Model and hyperparameters.} A weighted average of five ridge regressions with intercept, configured as (amplitude; inner power, outer power, run-in scale, regularization base, weight): difference; 3, 1/4, 10, 3, 1/4. Difference; 2.5, 1/3, 10, 3, 1/4. Projection; 2.5, 1/3, 10, 3, 1/4. Minor axis; 3, 1/4, 10, 3, 1/8. Minor axis; 2, 1/2, 5, 1, 1/8.

\noindent\textbf{Training and validation.} Sort by cutter and cut and join the development labels. Hold out whole cutters for validation, infer in causal order, and compare configurations and blends by the cutter-averaged MAE. In each fold, compute the input mean and standard deviation on the training part only; the regularization strength is the base times the number of training samples divided by 315. Finally, refit on all development samples and fold the standardization into the coefficients and intercept.

\noindent\textbf{Inference and output.} When cuts are skipped, fill the missing accumulation with the mean inner-power amplitude of the neighboring endpoints already seen; do not backfill on the first call. Reuse the output for a repeated cut; reset when cut numbers do not increase, and across cutters. Non-finite predictions fall back to the previous value, and on the first call to the mean first-cut label of the development data. Clip to $[0, 1000]$ and take the maximum with the previous prediction. Output finite predictions in the original order and identifiers.

\noindent\textbf{Compute.} Feature extraction for training uses four processes; inference uses one CPU and no GPU, with reading cost linear in record length and a constant history state; about 1\,s to start and 0.1\,s per cut, with peak memory of 256\,MB.
\par}

\begin{table*}[!t]
\centering
\caption{The 20 published methods that the strongest agent (GPT-6 Astra) implemented from the paper, with the score $S$ of each method and of Astra's own design on the same fold(s), from a run without a plan made at the same time as the implementations rather than the main run of Fig.~\ref{fig:scores}. Method $S$ is the score of the faithful implementation; for Paderborn it is the mean over the three folds, and Astra's own design is also averaged over them. Bold: the implementation beat Astra's design on its fold; $^\dagger$: on at least one of the three Paderborn folds. The last column counts the implementations (paper $\times$ fold) that beat Astra's design; ties do not count. Selection: on each dataset, the IEEE and MDPI papers since 2019 whose title or abstract names the dataset (OpenAlex, 17 October 2026), ranked by citations; after screening for method papers with supervised training on labelled data (no reviews, domain adaptation, few-shot, hardware, and so on, and no paper whose PDF could not be obtained automatically, which excluded MDPI copies), the three most cited, plus the two most cited since 2024. For N-CMAPSS, three papers come from a recent reproduction collection~\cite{theiler2026paper} (two of them in non-IEEE venues) and two from the same rule. Years are publication years of the issue. Scores are single runs; C-MAPSS, N-CMAPSS and Tennessee Eastman have one fold each.}
\label{tab:literature}
\footnotesize
\begin{tabular}{@{}llcrrr@{}}
\toprule
Dataset & Method & Year & $S$ (method) & $S$ (Astra's design) & Beat \\
\midrule
\multirow{5}{*}{C-MAPSS (FD004)} & Dual-task LSTM~\cite{miao2019dualtask} & 2019 & 21.0 & \multirow{5}{*}{44.7} & 0/1 \\
 & DA-TCN~\cite{song2021datcn} & 2021 & 23.4 &  & 0/1 \\
 & Two-stream Bi-LSTM~\cite{jin2022bilstm2stream} & 2022 & 21.4 &  & 0/1 \\
 & DSFormer~\cite{li2024dsformer} & 2024 & 9.4 &  & 0/1 \\
 & MSBLS~\cite{xu2024msbls} & 2024 & 9.3 &  & 0/1 \\
\midrule
\multirow{5}{*}{N-CMAPSS} & CART-Net~\cite{zhang2025cartnet} & 2025 & \textbf{43.2} & \multirow{5}{*}{38.1} & 1/1 \\
 & Failure prediction + RUL (LSTM)~\cite{cohen2023turbofan} & 2023 & 9.3 &  & 0/1 \\
 & Condition-indicator pipeline~\cite{graves2023turbofan} & 2023 & 27.6 &  & 0/1 \\
 & MR-LSTM~\cite{xu2024mrlstm} & 2024 & \textbf{48.3} &  & 1/1 \\
 & CDSG (graph NN)~\cite{wang2023cdsg} & 2023 & 30.6 &  & 0/1 \\
\midrule
\multirow{5}{*}{Paderborn} & Three-stage fusion~\cite{wang2021tim3stage} & 2021 & 32.6$^\dagger$ & \multirow{5}{*}{60.4} & 1/3 \\
 & LiteFormer~\cite{sun2024liteformer} & 2024 & 21.0 &  & 0/3 \\
 & MABFN~\cite{wang2023mabfn} & 2023 & 34.3 &  & 0/3 \\
 & CAFormer~\cite{tang2025caformer} & 2025 & 3.6 &  & 0/3 \\
 & MSHFFN~\cite{wang2025mshffn} & 2025 & 43.1 &  & 0/3 \\
\midrule
\multirow{5}{*}{Tennessee Eastman} & Enhanced RF (static/dynamic nodes)~\cite{chai2020erf} & 2020 & 0.0 & \multirow{5}{*}{30.1} & 0/1 \\
 & SKLFDA~\cite{zhong2020sklfda} & 2020 & 13.7 &  & 0/1 \\
 & MRCNN-LSTM fusion~\cite{liu2023mrcnnlstm} & 2023 & \textbf{46.1} &  & 1/1 \\
 & MRCC-Transformer~\cite{chen2025mrcctransformer} & 2025 & \textbf{47.5} &  & 1/1 \\
 & GNN, trainable adjacency~\cite{kovalenko2024gnnadj} & 2024 & \textbf{44.6} &  & 1/1 \\
\bottomrule
\end{tabular}
\end{table*}

\section{Models and Dataset Classes}
\label{app:models}
Table~\ref{tab:models} lists the models we evaluated; the providers' model snapshots were not recorded. Classed as less known, after the runs, were AHU, CARE, CFRP, MATWI, the NASA randomized battery, PHM Europe 2020, SCANIA, and Petrobras 3W; the other 16 datasets are standard benchmarks with many published results.

\begin{table}[!htbp]
\centering
\caption{Models evaluated. Open-weight models were served with llama.cpp.}
\label{tab:models}
\footnotesize
\resizebox{\columnwidth}{!}{%
\begin{tabular}{@{}lll@{}}
\toprule
Model & Access & Harness, runs \\
\midrule
GPT-6 Astra & API & Codex 0.156.1$^{a}$, Sep--Oct 2026 \\
GPT-6 Luna & API & Codex 0.156.1$^{a}$, Sep--Oct 2026 \\
Kimi K3 & API & Codex 0.156.1, Sep 2026 \\
DeepSeek V4.1-Flash & API & Codex 0.156.1, Sep 2026 \\
Qwen3.6-35B-A3B & 4-bit (IQ4\_XS) & OpenCode 1.19.0, Oct 2026 \\
Gemma 4 26B-A4B & 4-bit (IQ4\_XS) & OpenCode 1.19.0, Oct 2026 \\
\bottomrule
\multicolumn{3}{@{}l}{$^{a}$Plan runs and their controls used OpenCode 1.19.0.}
\end{tabular}}
\end{table}

\section{Toolkit Trials}
\label{app:toolkits}
Table~\ref{tab:toolkits} lists the fifteen agent--toolkit combinations of Section~\ref{sec:levers}. Each was run on the datasets the toolkit applies to and compared with the same agent, in the same harness, without the toolkit.

\begin{table}[!htbp]
\centering
\caption{The fifteen toolkit trials behind Fig.~\ref{fig:levers}a: seven toolkits, each given to one or more frontier agents, through Codex from the start of the run unless marked. Control: the agent's own runs without the toolkit in the same harness (for Codex, the mean of two runs). $\Delta S$ is the paired change over the datasets covered, with 95\% intervals from a bootstrap over datasets.}
\label{tab:toolkits}
\footnotesize
\resizebox{\columnwidth}{!}{%
\begin{tabular}{@{}llcc@{}}
\toprule
Toolkit & Agent & Datasets (folds) & $\Delta S$ \\
\midrule
AWS predictive-maintenance skill & Luna & 24 (58) & $-$2.4 [$-$8.7, +3.5] \\
tsfresh features & Luna & 24 (58) & +1.2 [$-$3.4, +5.7] \\
Predictive-maintenance server & Luna & 9 (25) & $-$1.0 [$-$9.7, +8.8] \\
PyOD & Luna & 4 (10) & +2.2 [$-$5.5, +12.1] \\
ProgPy & Luna & 15 (35) & $-$0.9 [$-$4.9, +3.2] \\
Own feature components & Luna & 24 (58) & +1.7 [$-$3.3, +7.4] \\
Own components and validation tools & Luna & 24 (58) & $-$0.7 [$-$5.7, +3.9] \\
AWS predictive-maintenance skill$^{a}$ & Luna & 24 (57) & $-$0.8 [$-$3.8, +2.3] \\
AWS predictive-maintenance skill & Astra & 24 (58) & $-$3.2 [$-$7.0, +0.2] \\
tsfresh features & Astra & 24 (58) & $-$4.8 [$-$9.2, $-$0.6] \\
Predictive-maintenance server & Astra & 9 (25) & +3.1 [$-$1.0, +8.3] \\
PyOD & Astra & 4 (10) & $-$2.3 [$-$9.5, +4.2] \\
ProgPy & Astra & 15 (35) & +1.2 [$-$4.2, +7.8] \\
Own feature components & Kimi & 24 (58) & $-$2.8 [$-$7.3, +2.0] \\
AWS predictive-maintenance skill$^{b}$ & Luna & 24 (58) & $-$0.8 [$-$5.5, +4.0] \\
\midrule
Mean of the fifteen & & & $-$0.7 \\
\bottomrule
\multicolumn{4}{@{}l}{$^{a}$Given after the agent's first submission. $^{b}$Through OpenCode; control is one OpenCode run.}
\end{tabular}}
\end{table}

\section{Alternative Fixed Baselines}
\label{app:baselines}
\begin{figure}[!htbp]
\centering
\includegraphics[width=\columnwidth]{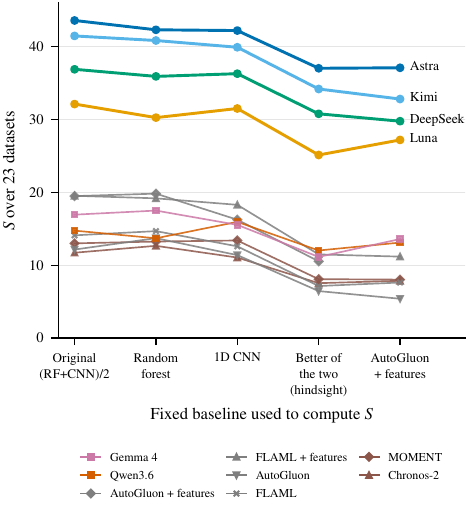}
\caption{Score $S$ over 23 datasets of every method under five choices of the fixed baseline. Stricter baselines lower every score, but the methods move together; the four frontier agents stay on top, in the same order. A method is not scored against itself.}
\label{fig:altbase}
\end{figure}

Every fold is scored against a fixed baseline, the average loss of a random forest and a small one-dimensional CNN, and on a few folds the CNN is weak, which raises every score. We therefore rescored the predictors every method delivered against other baselines, changing only the denominator: either of the two models alone; the better of the two baseline models on each fold, chosen with hindsight; and the strongest AutoML pipeline, AutoGluon with the feature bank. A method is not scored against itself. Fig.~\ref{fig:altbase} and Table~\ref{tab:altbase} report $S$ over the 23 datasets that every method covers; AutoML was not run on MIMII. A stricter baseline lowers every score, but the methods move together, and the four frontier agents keep their order under all five baselines. Seven further baselines, AutoGluon without features, FLAML with and without features, the two foundation models, a constant prediction, and Qwen3.6, give the same order in every case but the last: Qwen3.6 fails on enough folds to leave 37 to score, and on those Kimi edges past Astra, 42.1 to 40.6. Under all twelve, the main comparisons of this paper keep their sign.

\begin{table}[!htbp]
\caption{Score $S$ over 23 datasets under five fixed baselines. Ranks of the frontier agents among the twelve methods in parentheses; ---: the method is itself the baseline. A constant prediction (the median or majority of the development labels) is a zero-information reference; over all 24 datasets it scores 3.9.}
\label{tab:altbase}
\centering
\footnotesize
\resizebox{\columnwidth}{!}{%
\begin{tabular}{@{}lccccc@{}}
\toprule
 & Original & Random & & Better of & AutoGluon \\
Method & (RF+CNN)/2 & forest & 1D CNN & the two & + features \\
\midrule
Astra & 43.6 (1) & 42.3 (1) & 42.2 (1) & 37.0 (1) & 37.1 (1) \\
Kimi & 41.5 (2) & 40.8 (2) & 39.9 (2) & 34.2 (2) & 32.8 (2) \\
DeepSeek & 36.9 (3) & 35.9 (3) & 36.3 (3) & 30.8 (3) & 29.8 (3) \\
Luna & 32.1 (4) & 30.3 (4) & 31.5 (4) & 25.1 (4) & 27.2 (4) \\
\midrule
Qwen3.6 & 14.7 & 13.7 & 15.9 & 12.0 & 13.1 \\
Gemma 4 & 16.9 & 17.5 & 15.5 & 11.1 & 13.6 \\
\midrule
AutoGluon & 12.1 & 13.7 & 11.4 & 6.4 & 5.4 \\
FLAML & 14.1 & 14.7 & 12.6 & 7.1 & 7.6 \\
AutoGluon + features & 19.5 & 19.8 & 16.2 & 10.5 & --- \\
FLAML + features & 19.5 & 19.2 & 18.3 & 11.5 & 11.2 \\
\midrule
MOMENT-1-base & 13.0 & 13.2 & 13.4 & 8.1 & 8.0 \\
Chronos-2 & 11.7 & 12.7 & 11.0 & 7.5 & 7.9 \\
\midrule
Constant prediction & 4.1 & 4.1 & 4.9 & 2.0 & 3.4 \\
\bottomrule
\end{tabular}}
\end{table}

\section{Published Methods Implemented}
\label{app:lit}
Table~\ref{tab:literature} lists the 20 papers whose methods the strongest agent implemented, with the score of each implementation and of the agent's own design on the same folds.

\FloatBarrier
\section*{Acknowledgment}
Part of the computations in this work, mainly the runs of local open-weight models, time-series foundation models, and AutoML, were performed using computational resources from the ``M\'esocentre'' computing center of Universit\'e Paris-Saclay, CentraleSup\'elec and \'Ecole Normale Sup\'erieure Paris-Saclay supported by CNRS and R\'egion \^Ile-de-France (\url{https://mesocentre.universite-paris-saclay.fr/}).

This paper practices what it argues. The ideas, the study design, the structure of the paper, its key arguments and formulations, and the interpretation of results are the authors' own. Under their direction, AI assistants (OpenAI GPT and Anthropic Claude models) implemented the benchmark infrastructure and analysis scripts, prepared figures, and helped turn the authors' outline into English prose and polish it. The assistants reviewed each other's code, and separate sessions audited the pipeline independently; none took part in the evaluated runs. The authors checked all reported numbers against the run records and take full responsibility for the content.

\bibliographystyle{IEEEtran}
\bibliography{refs}

\end{document}